\documentclass[logo]{deepmind}

\title{Revisiting Gaussian mixture critics in off-policy reinforcement learning: a sample-based approach}

\author[1*]{Bobak Shahriari}
\author[1*]{Abbas Abdolmaleki}
\author[1]{Arunkumar Byravan}
\author[1]{Abe Friesen}
\author[1]{Siqi Liu}
\author[1]{Jost Tobias Springenberg}
\author[1]{Nicolas Heess}
\author[1]{Matthew W. Hoffman}
\author[1]{Martin Riedmiller}
\affil[1]{DeepMind}
\affil[*]{corresponding authors; equal contribution}

\usepackage{color}
\usepackage[normalem]{ulem}
\usepackage{hyperref}
\usepackage{url}
\usepackage{amsmath}
\usepackage{nicefrac}
\usepackage{natbib}
\setcitestyle{square, numbers}
\usepackage{graphicx}
\usepackage{subcaption}
\usepackage{algpseudocode}
\usepackage{algorithm}
\usepackage{balance}
\usepackage[colorinlistoftodos]{todonotes}
\usepackage[T1]{fontenc}
\usepackage{booktabs}       % professional-quality tables
\usepackage{amsfonts}       % blackboard math symbols
\usepackage{microtype}      % microtypography
\usepackage{float}
\newfloat{algorithm}{t}{lop}
\usepackage[english]{babel}
\usepackage[utf8]{inputenc}
\usepackage{amssymb}
\usepackage{amsthm}
\usepackage{wrapfig}

\DeclareUrlCommand\ULurl{%
  \renewcommand\UrlFont{\ttfamily\color{blue}}%
  \renewcommand\UrlLeft{\uline\bgroup}%
  \renewcommand\UrlRight{\egroup}}

\usepackage{mwhmacros}

\usepackage{hyperref}
\usepackage{url}

\begin{abstract}
Actor-critic algorithms that make use of distributional policy evaluation have frequently been shown to
outperform their non-distributional counterparts on many challenging control tasks. Examples of this behavior 
include the D4PG and DMPO algorithms as compared to DDPG and MPO, respectively~\citep{d4pg,hoffman2020acme}. However, both agents
rely on the C51 critic for value estimation. One major drawback of the C51 approach is its
requirement of prior knowledge about the minimum and maximum values a policy can
attain as well as the number of bins used, which fixes the resolution of the
distributional estimate. While the DeepMind control suite of tasks utilizes standardized rewards and
episode lengths, thus enabling the entire suite to be solved with a single
setting of these hyperparameters, this is often not the case.
This paper revisits a natural alternative that removes this requirement, namely a mixture of Gaussians, and a simple sample-based loss function to train it in an off-policy regime.
We empirically evaluate its performance on a broad range of continuous control tasks and demonstrate that it eliminates the need for these distributional hyperparameters and achieves state-of-the-art performance on a variety of challenging tasks (e.g. the humanoid, dog, quadruped, and manipulator domains).
Finally we provide an implementation in the Acme agent repository.
\end{abstract}

\begin{document}
\maketitle

\section{Introduction}

The field of Reinforcement Learning (RL) formalizes the study and design of agents which 
interact with their environment and make observations \citep{Sutton1998}. A typical RL agent is one which optimizes 
its behavior in order to maximize its expected return---i.e.\ the expected (discounted) sum of future rewards provided 
by the environment. However, for many algorithms a crucial prerequisite to this optimization process is the ability
to \emph{predict} the agent's expected return. As a result a key building block of such agents is their value function,
i.e.\ a learned function which maps from the current state of the environment to the return expected under future interaction.
This function can be used in an iterative fashion by first improving its predictions and using the updated
values to improve the agent's behavior; this process can be repeated until the agent's behavior converges.

Classically, value-based RL algorithms rely on a function which makes point estimate predictions of the expected return. While
algorithms making use of more complex predictions or higher moments do exist \citep{morimura2012parametric,prasanth2013actor,tamar2016learning}, they are far from the norm.
Recently, though, the work of \citep{BellemareDM17} has revived interest in this domain by proposing a mechanism to estimate
a distribution over values rather just a single scalar value. By making use of this distributional approach to value estimation \citeauthor{BellemareDM17}
was able to achieve what was at the time, state-of-the-art performance on the Atari 2600 benchmark suite. These techniques were later extended
to continuous action-space Actor-Critic algorithms in the form
of D4PG and DMPO \citep{d4pg,hoffman2020acme} achieving state-of-the-art performance in many continuous control tasks. 

By design the distribution over values has an expectation equivalent to the standard value function, a key point that allows these
methods to easily integrate with many mechanisms for policy improvement. However, the precise form of this distribution plays a key role in
how it is updated based on incoming reward observations. In the case of the methods presented thus far this distribution 
is represented by a categorical distribution defined at a regular grid of atoms over the space of expected returns---i.e.\ the real line.
This use of a discrete distribution over the space of returns has two potential downsides.
First, returns are inherently continuous and as a result there is some loss of representation due to the use of this discretized distribution.
Although at first
glance this may seem problematic it is still possible to increase the number of atoms in this representation. While such an approach
does increase the complexity of the
algorithm, it does so only linearly since the returns that must be predicted are a scalar quantity. 
More importantly, however, the categorical distribution in question has bounded support that must be known ahead of time.

While it is possible to learn the support of the return distribution \citep[see e.g.][]{dabney18iqn} an
alternative is to make use of a distribution which is naturally able to shift its support.
In this work we revisit
an alternative distributional value function parameterized by a mixture of Gaussians (MoG)
and the simple sample-based approach to optimize this object, which was first proposed in~\citep[][appendix A--C]{d4pg}. Compared to its categorical counterpart, the MoG parameterization has
received relatively little scrutiny, a fact we attribute
primarily to previously inferior results. To address this we also provide a number of
experiments to examine the performance of our approach when used
within a modern off-policy actor-critic algorithm. Our results show that this technique is able to achieve state-of-the-art performance on a number of hard continuous
control tasks.
Finally, we provide some preliminary analysis~(Sections~\ref{sec:analysis} and~\ref{sec:sensitivity}) to help understand what made this implementation successful
compared to prior attempts~\citep{d4pg}.

Overall, the primary contribution of this work is an effective combination of a mixture-of-Gaussians
value distribution with the MPO algorithm for policy optimization that we found particularly effective
for continuous control domains. 
Our results in \ref{sec:analysis} closely examines the reasons why this approach is successful, 
and may prove useful for further development of distributional algorithms.
Finally, while we did find this algorithm to be fairly robust, we did find it to be somewhat sensitive to the choice of initial scale of the Gaussian components; see \ref{sec:sensitivity}. 

\subsection{Related work}

The approach described in this paper is most related to two particularly relevant lines of work.
The first line is that of~\citep{BellemareDM17}, whose work
renewed recent interest in distributional RL as a whole as well as presenting both
both initial theoretical analysis and impressive empirical Atari results. The critic introduced by \citeauthor{BellemareDM17} 
outputs a mixture of fixed delta masses, an approach that has influenced much of the following distributional RL work and 
that we will refer to as a categorical critic in this report (or C51 when using 51 atoms as prescribed by \citeauthor{BellemareDM17}).
The second related line of work is that of~\citet{d4pg} and~\citet{hoffman2020acme}, who have successfully
combined categorical critic with the DDPG and MPO policy optimization approaches, 
respectively, in order to improve on the state-of-the-art in continuous control.

While the works of \citet{BellemareDM17} and \citet{d4pg} focus on categorical value distributions, 
the latter work also describes and experiments with a mixture-of-Gaussians distribution. This
earlier work, however, found the MoG approach to underperform compared with that of the 
categorical distribution and left its use relatively unexplored. Much earlier work of 
\citet{morimura2012parametric} uses a single Gaussian for its return distribution and minimizes
the KL divergence in a similar way to this paper. However this earlier work also uses much simpler policies
which do not make use of modern deep neural networks and as a result are not directly comparable.
More recently, the works of \citet{gmac} and \citet{mogq} have returned the the question of a 
mixture-of-Gaussians distribution. The closest of these to our work is the approach of $\text{SR}(\lambda)$ 
presented in \citep{gmac} which differs in two primary ways. The first is in the choice of the divergence metric 
between distributions; unlike our approach $\text{SR}(\lambda)$ makes use of the Cram\'er metric rather than the 
KL divergence. The second, and perhaps bigger difference, is the choice of underlying policy optimization metric 
in which $\text{SR}(\lambda)$ makes use of PPO (whereas we use MPO)---as a result this work need only estimate 
the value function $V$
and is somewhat more on-policy than the approach described in this work. Although interesting, we leave an 
in-depth comparison to $\text{SR}(\lambda)$ for future work. Similarly the MOG-Q algorithm introduced in 
\citep{mogq} differs greatest in its choice of policy optimization algorithm---DQN in this case. The use of DQN 
however leads to a greater difference in its loss function and requires the authors to introduce their JTD loss. 
Additionally this approach makes MOG-Q less applicable to the continuous control examples focused on by this 
work.

In terms of relevant theoretical results, Lemma~3 of~\citep{BellemareDM17} shows that the
distributional Bellman operator is a
$\gamma$-contraction in the supremum-Wasserstein metric.
In this report, however, we do not benefit from Lemma~3, as we do not minimize the 
appropriate error metric. This should be considered in future work, perhaps capitalizing
on the Cram\'er distance and Proposition~2 from~\citep{rowland2018analysis}, which
demotes the result to a $\sqrt{\gamma}$-contraction in the supremum-Cram\'er metric.
Since in the actor-critic setting we have separate policy evaluation and improvement 
subroutines, the purpose of the former is simply to get an estimate of $Q_\pi$, therefore
we are not interested in the distributional Bellman optimality operator
from~\citep{BellemareDM17} and we are not affected by the negative results in Lemma~4 and
Theorem~1 therein, which proves that the optimality operator is not a contraction.

There are also several works that are left out of the scope of this report which we should nevertheless mention
as additional comparisons and potential future work.
For instance recent quantile-based work by~\citet{dabney18iqn, dabney2018qr},
which also removes the hyperparameter requirements of C51; these are more complex algorithms
than the one studied in this manuscript and a full comparison is left for future work. Concurrently to the present work,
\citet{nguyen2021distributional} have explored similar sample-based approaches by leveraging the maximum mean
discrepancy~\citep[MMD;][]{gretton2012kernel}; this may be a promising alternative although it brings with it
a kernel function as a hyperparameter.
Finally, somewhat related albeit non-distributional is the PopArt
approach of~\citet{van2016learning}, which applies an adaptive standardization transformation to the value
targets; a similar approach may be useful to remove the vmin/vmax hyperparameters, although the fixed
number of delta masses of C51 would remain and would bound the resolution of values that can be considered.

\section{Background and notation}
\label{sec:problem}

We adopt the standard formalization of reinforcement learning (RL) as a Markov decision process (MDP)
and much of the notation from~\citet{rowland2018analysis}.
In particular we denote with $\calS$ and $\calA$ the spaces of states
and actions respectively; a transition distribution $p(R_t,S_{t+1}\,|\, S_t, A_t)$ over
immediate rewards and next states conditioned on a given state-action pair; a discount factor $\gamma \in [0, 1)$;
and we denote an agent's policy with $\pi(A_t\,|\,S_t)$.
A trajectory of states, actions, and rewards will be denoted with $(S_t, A_t, R_t)_{t\geq0}$ where we have
used capital letters to emphasize the fact that these are random variables.
To generate a trajectory the environment starts from an initial state $s_0 \in \calS$ and for every subsequent timestep
$t \geq 0$, an agent produces an action $a_t \sim \pi(\cdot\,|\,S_t=s_t)$ sampled from its policy.
The environment then produces an immediate reward and next state according to the transition distribution
$(r_t, s_{t+1}) \sim p(\cdot, \cdot \,|\,S_t=s_t, A_t=a_t)$. 
We can then define the random variable $Z_\pi^{(s, a)}$ representing the sum of $\gamma$-discounted rewards, conditioned on an 
initial state and action $(s, a)$:
\begin{equation}
    Z_\pi^{(s, a)} = \sum_{t=0}^\infty \gamma^t R_t \ \bigg|\ S_0 = s, A_0 = a,
\end{equation}
where the implicit dependence on the policy $\pi$ is indicated as a subscript on $Z$.
Hereafter we refer to this quantity as the return of policy $\pi$. Classical off-policy
evaluation methods are interested in evaluating a policy's expected return given an
arbitrary initial state-action pair $(s, a)$, a quantity known as the state-action value
function 
    $Q_\pi(s, a) = \E Z_\pi^{(s, a)}$
where the expectation is taken with respect to the trajectory distribution induced by
the policy $\pi$ and the unknown transition distribution $p$.
In distributional RL, we are interested in the distribution of the conditional random
variable $Z_\pi^{(s, a)}$ rather than only its first moment $Q_\pi(s,a)$.
Therefore, in a slight abuse of notation, we will also let $Z_{\pi}^{(s,a)}$ denote the 
probability density function of the eponymous random
variable as the symbol's meaning should always be clear from context.

In this report, we focus on off-policy actor-critic algorithms. These are algorithms that
combine a policy $\pi$ with a critic $Q_\pi$, alternating between (a) training the
critic to evaluate the current policy, i.e., \emph{policy evaluation}, and (b) training
the policy to shift its probability mass towards higher quality actions (according to 
the critic), i.e., \emph{policy improvement}.
Since this study is focused on distributional policy evaluation, for most of this
report we fix the policy improvement algorithm to be maximum \emph{a posteriori} policy 
optimization~\citep[MPO;][]{abdolmaleki2018relative,abdolmaleki2018maximum}, although we
show in Figure~\ref{fig:benchmark-ddpg} that our findings generalize to deterministic
policy gradient algorithms~\citep{silver2014deterministic,d4pg}. 
In both settings the policy improvement
steps target the expectation of the distributional critic estimate, i.e.\
$Q_\pi = \E Z_\pi$ as produced by the policy evaluation step.

\begin{figure}
    \centering
    \includegraphics[width=0.8\textwidth, trim=0 50 0 50, clip]{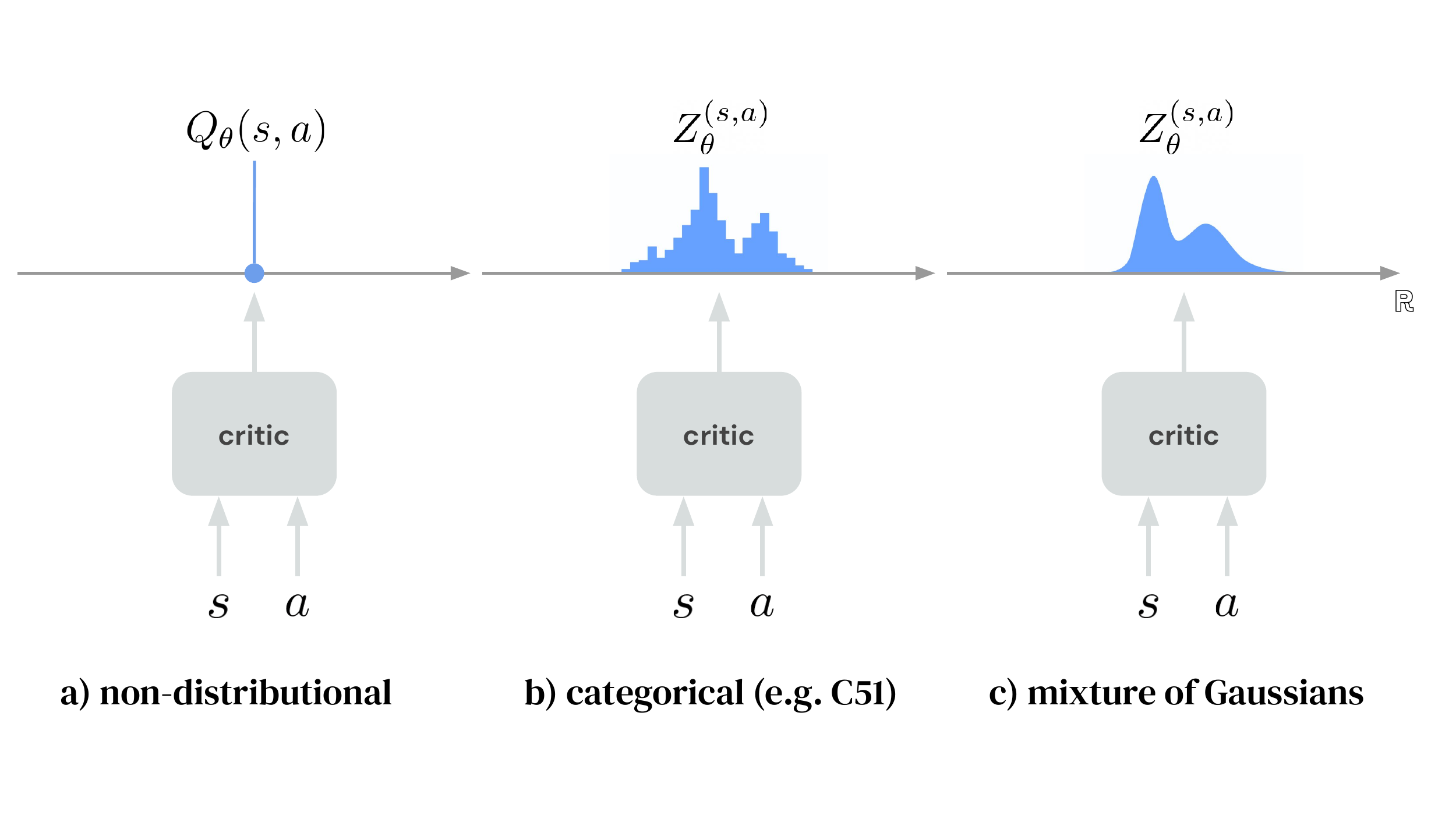}
    \caption{
        Diagram depicting the three different kinds of critics considered in this report.}
\end{figure}

In order to discuss policy evaluation we must first introduce the distributional Bellman operator.
Following \citep{BellemareDM17} for an arbitrary scalar random variable $Z$ which can be conditioned 
on state-action pairs, the operator in question can be written as
\begin{align}
    (\calT^\pi Z)^{(s, a)}
        &= \mathbb{E}_{r, s' \sim p} \bigg[ \mathbb{E}_{a' \sim \pi} 
        \big[r + \gamma Z^{(s',a')}\big]\ \big{|}\ s, a \bigg].
    \label{eq:distributional-bellman-operator}
\end{align}
With this operator in hand, and 
analogously to non-distributional Q-learning, the distribution function $Z_\pi$ satisfies
the distributional Bellman equation, i.e.\ that the distribution is a fixed point under the distributional
Bellman operator. For any state and action this can be written as
\begin{align}
    Z_\pi^{(s,a)} = (\calT^\pi Z_\pi)^{(s,a)}.
\end{align}
This can then be used to perform policy evaluation by starting from an arbitrary estimate $Z$ and using the Bellman
operator to move this estimate towards its fixed point.

\section{Implementation}

\algnewcommand\algorithmicforeach{\textbf{for each}}
\algdef{S}[FOR]{ForEach}[1]{\algorithmicforeach\ #1\ \algorithmicdo}

The algorithm described in this work can be summarized as a combination of the following key ideas:

\begin{itemize}
    \setlength\itemsep{1em}
    \item \textbf{Parameterize the distributions.} We use a mixture of Gaussians (MoG) to
    parameterize the distributions $Z_\pi$ because it is very expressive, it can be sampled from,
    and has a differentiable log-probability density function.
    It also has infinite support, obviating the need for projections onto a fixed support, unlike its
    categorical counterparts~\citep{BellemareDM17,d4pg}.
    
    \item \textbf{Neural network parameterization.} We use a neural network, denoted $Z_\theta$, to
    approximate $Z_\pi^{(s,a)}$ and, as is common, we also use a target network $Z_{\bar\theta}$ to
    estimate bootstrap targets to stabilize Q-learning. In this work, the target network is
    periodically updated but we do not anticipate any issues with using incremental target updates.
    
    \item \textbf{Distributional loss.} A distributional loss (distance or
    divergence) $d(Z, Z')$ is necessary to to minimize the distributional
    Bellman error which we can write as $d(\calT^\pi Z_{\bar\theta}, Z_\theta)$, i.e.\
    the difference between the return distribution before and after applying the distributional Bellman operator.
    In practice we use the cross-entropy loss $H$, which
    although it does not guarantee a fixed point 
    this loss works well in practice; we leave the consideration of other losses for future work.
        
    \item \textbf{Sample-based approximation.} We use Monte Carlo approximation for all
    integrals required to compute the distributional Bellman 
    operator~\eqref{eq:distributional-bellman-operator}, creating a fully sample-based
    empirical critic loss.
\end{itemize}
Combining all of these components, we arrive at the following expression for our distributional loss:
\begin{align}
    L(\theta)
    = -\E_{s, a} H \left[
        \calT^\pi Z_{\bar\theta}^{(s, a)}, Z_\theta^{(s, a)}
    \right]
    &= -\E_{s,a} H \left[\E_{r,s'} \E_{a' \sim \pi} \big[r + \gamma Z_{\bar\theta}^{(s',a')}\big], Z_\theta^{(s, a)}
    \right]
\intertext{
    where the expectation over $(s,a)$ is taken over a replay buffer as usual in lieu of the
    stationary distribution of the policy $\pi$.
    As we do not have access to the environment's transition kernel or reward function,
    for every $(s, a)$ we naturally only have the observed subsequent $(r, s')$ as samples available
    for Monte Carlo integration. Omitting the indices on the $B$ minibatch samples from replay,
    we obtain the following loss
}
    &= -\frac1B \sum_{s,a,r,s'} H \left[\E_{a' \sim \pi} 
        \big[r + \gamma Z_{\bar\theta}^{(s',a')}\big], Z_\theta^{(s, a)}
    \right]
    = -\frac1B \sum_{s,a,r,s'} \ell(\theta; s, a, r, s'),
\end{align}
We have implicitly defined a per-transition loss function $\ell$ as the expected cross-entropy
Bellman error over (unobserved) next actions according to the policy $\pi$.
Therefore we have two more integrals left to approximate, the outer one over the real line representing
possible returns, this is the one hidden by the cross-entropy symbol $H$; and the inner one over the
policy we are bootstrapping with respect to.
Again we estimate these via Monte Carlo, generating sampled returns by transforming samples from
$Z_{\bar\theta}^{(s', a')}$ to get the final expression for our empirical critic loss:
\begin{align}
    \ell(\theta; s, a, r, s')
    &\approx -\frac1{NM} \sum_{i, j=1}^{N, M} \log Z_\theta^{(s, a)}\left(r + \gamma z'^{(i,j)}\right),
    \quad \text{where} \quad
    \begin{cases}
    \ z'^{(i, j)} \sim Z_{\bar\theta}^{(s', a'^{(j)})}, \text{and}
    \\
    \ a'^{(j)} \sim \pi(\cdot | s').
    \end{cases}
\end{align}
In practice we often bootstrap with respect to the greedy policy such that $M=1$ and $a'$ is the mode
of the stochastic policy $\pi(\cdot | s')$.
Notice that the only requirements for this approximate loss are that (i) we can sample realizations
of the random variable $Z_{\bar\theta}$ and (ii) we can compute differentiable log-probabilities.
This means that while in this report we focus on a mixture-of-Gaussians parameterization of the
distributional Q-function, this loss can be used more broadly.

\begin{algorithm}
    \caption{Sample-based distributional loss}\label{alg:sample-based-distributional-loss}
    \begin{algorithmic}[1]
        \Require Mini-batch $\calB$ of size $B$, online and target distributional networks $Z_\theta$ and $Z_{\bar\theta}$, discount $\gamma$, number of action samples $M$, number of return samples $N$
        \ForEach{$(s, a, r, s')$ in $\calB$}
            \State $a'^{(j)} \sim \pi(\cdot | s')$ for $j=1\dots M$  \Comment{Alternative: use mode of $\pi(\cdot|s')$ here to evaluate greedy policy.}
            \State $z'^{(i,j)} \sim Z_{\bar\theta}^{(s',a'^{(j)})}$ for $i,j = 1\dots N, M$ \Comment{Sample returns from target critic network's distributional output.}
            \State $\ell(\theta; s, a, r, s') \gets -\frac1{NM} \sum_{i,j} \log Z_\theta^{(s,a)}(r + \gamma z'^{(i,j)})$
        \EndFor\\
        \Return $\frac1B \sum_{\calB} \ell(\theta; s, a, r, s')$
    \end{algorithmic}
\end{algorithm}

\section{Experiments}

\begin{figure}
    \centering
    \includegraphics[height=0.70\textwidth, angle=270]{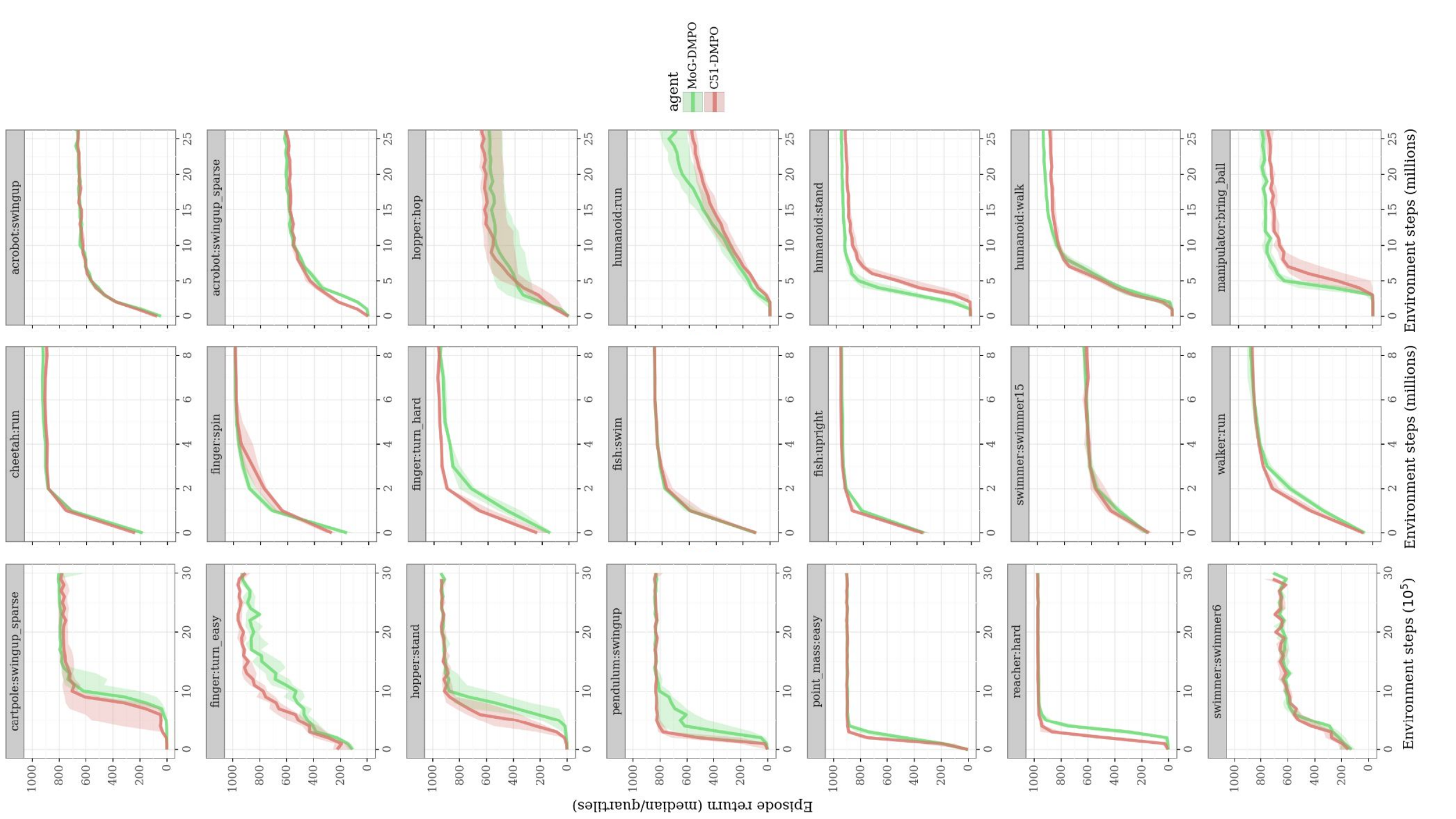}
    \caption{
        Benchmark runs on control suite tasks,
        showing median and quartiles across 10 seeds.
        Comparing MPO with a C51 (red) and a sample-based MoG (green) critic.
        On most tasks, the final performance of the MoG matches that of the 
        state-of-the-art C51, which corresponds the DMPO from~\citep{hoffman2020acme}.
        On the most challenging tasks in the humanoid and manipulator
        domains, the MoG critic provides a significant performance boost.
    }
    \label{fig:benchmark-control-suite}
\end{figure}

We compared the Mixture of Gaussian critic with 5 components to two other baselines:
the C51 critic and the non-distributional critic. In all cases we fixed the policy 
optimizer to MPO with an identical configuration. All three critics shared the exact same
network architecture and hyperparameters: batch sizes, learning rates, etc.
Note that this meant adding a 51-unit layer to the non-distributional critic for a fair 
comparison. Please refer to the released agent source code for the default hyperparameters used\footnote{
    \texttt{https://github.com/deepmind/acme/blob/master/acme/agents/tf/\{mpo|dmpo|mog\_mpo\}},
    where \texttt{mpo} refers to the agent with a non-distributional critic in the following results
    and similarly \texttt{dmpo} refers to the MPO agent with a C51 critic.
}.

\begin{figure}
    \centering
    \includegraphics[width=0.7\textwidth]{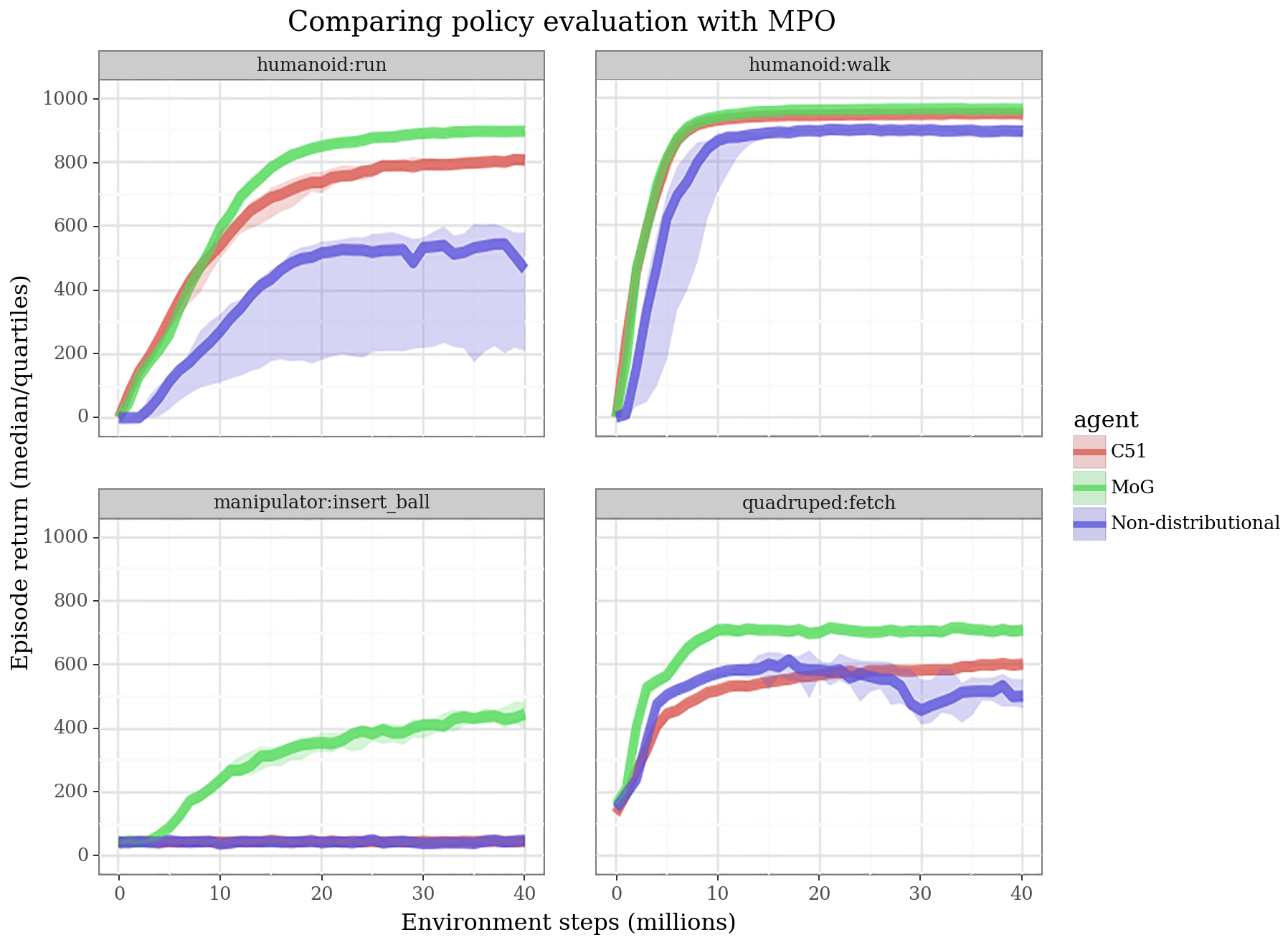}
    \caption{
        Benchmark runs on a selection of control suite tasks,
        showing median and quartiles across 10 seeds.
        Comparing MPO policy improvement with various policy evaluation approaches.
        The mixture of Gaussian evaluation trained with the sample-based method consistently outperforms the alternatives.
    }
    \label{fig:benchmark-selected}
\end{figure}

\begin{figure}
    \centering
    \includegraphics[width=0.8\textwidth]{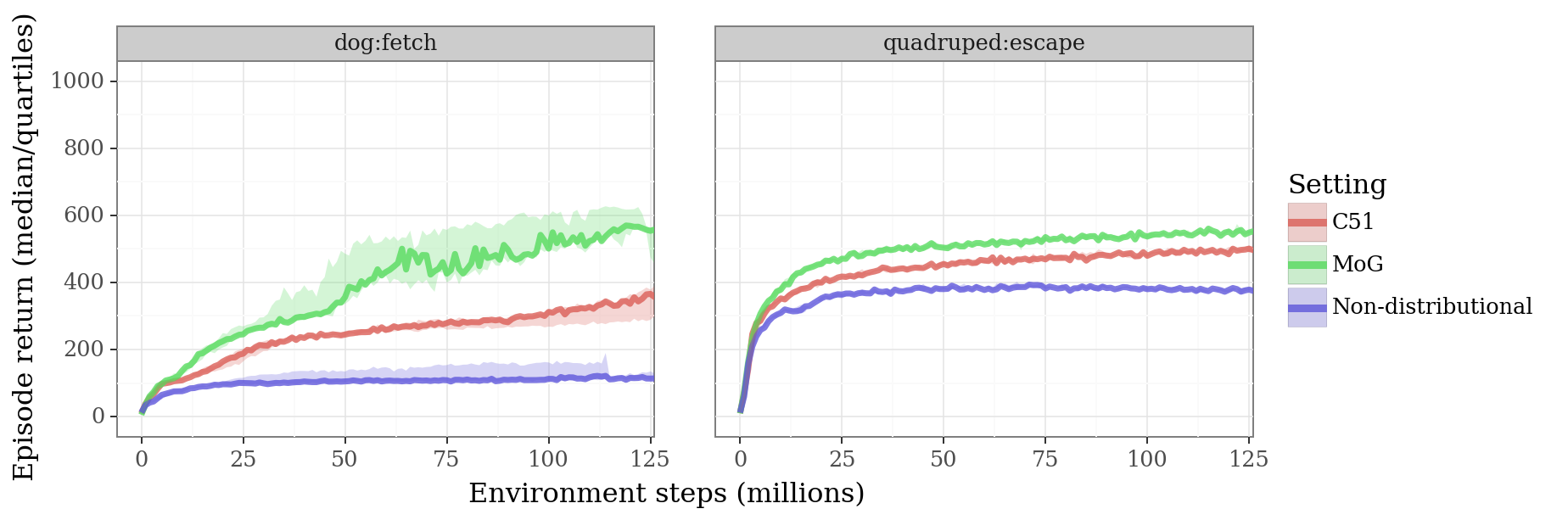}
    \caption{
        Benchmark runs on more challenging control suite tasks,
        showing median and quartiles across 10 seeds.
        Comparing MPO policy improvement with various policy evaluation approaches.
        The mixture of Gaussian evaluation trained with the sample-based method consistently outperforms the alternatives.
    }
    \label{fig:benchmark-extras}
\end{figure}

\begin{figure}
    \centering
    \includegraphics[width=0.8\textwidth]{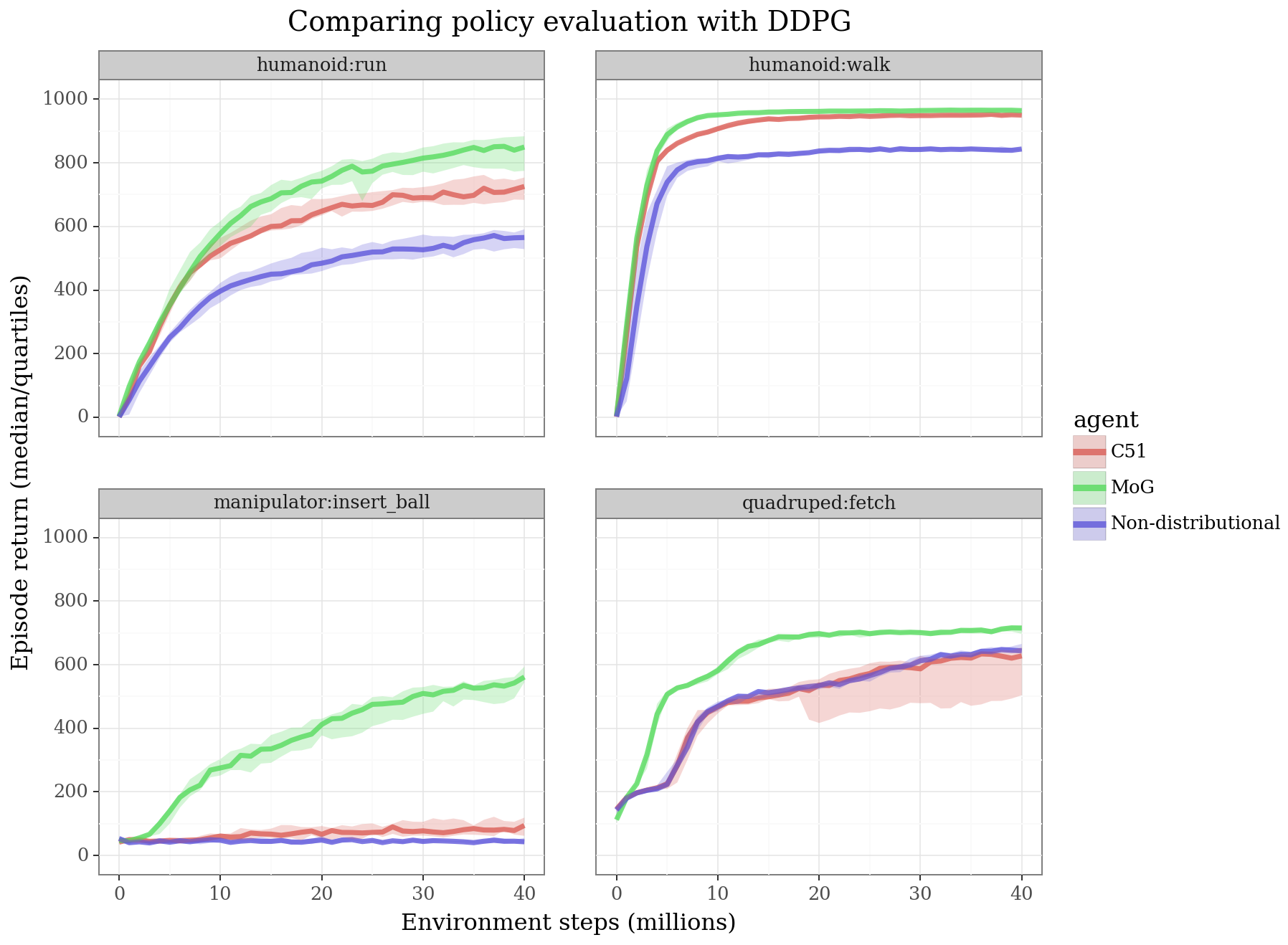}
    \caption{
        Benchmark runs on a selection of control suite tasks.
        Comparing DDPG policy improvement with various policy evaluation approaches;
        showing median and quartiles over 5 seeds.
        This figure shows that our findings generalize to other policy improvement algorithms.
        Note that the C51 learning curve corresponds exactly to the D4PG agent.
    }
    \label{fig:benchmark-ddpg}
\end{figure}

On 25 control suite environments the MoG critic achieves similar results on simpler tasks
when compared to C51 as shown in Figure~\ref{fig:benchmark-control-suite}.
However we observed much better performance for more difficult tasks,
shown in Figure~\ref{fig:benchmark-selected} and Figure~\ref{fig:benchmark-extras}.
Note that use of C51 in this setting already provides a strong baseline as shown by~\citet{hoffman2020acme} under the name of DMPO.
In comparison, the MoG variant overall achieves similar or better performance across the board, while not
having the limitations of C51, e.g.\ requiring the knowledge of minimum and maximum
Q-values.
Additionally the exact same comparison with DDPG instead of MPO revealed the same trends shown
in Figure~\ref{fig:benchmark-ddpg}, confirming that the gains are agnostic to the policy
optimization approach.

\section{Detailed analysis of the MoG distribution}
\label{sec:analysis}

\begin{figure}
    \centering
    \includegraphics[width=\textwidth, trim=0 40 0 40, clip]{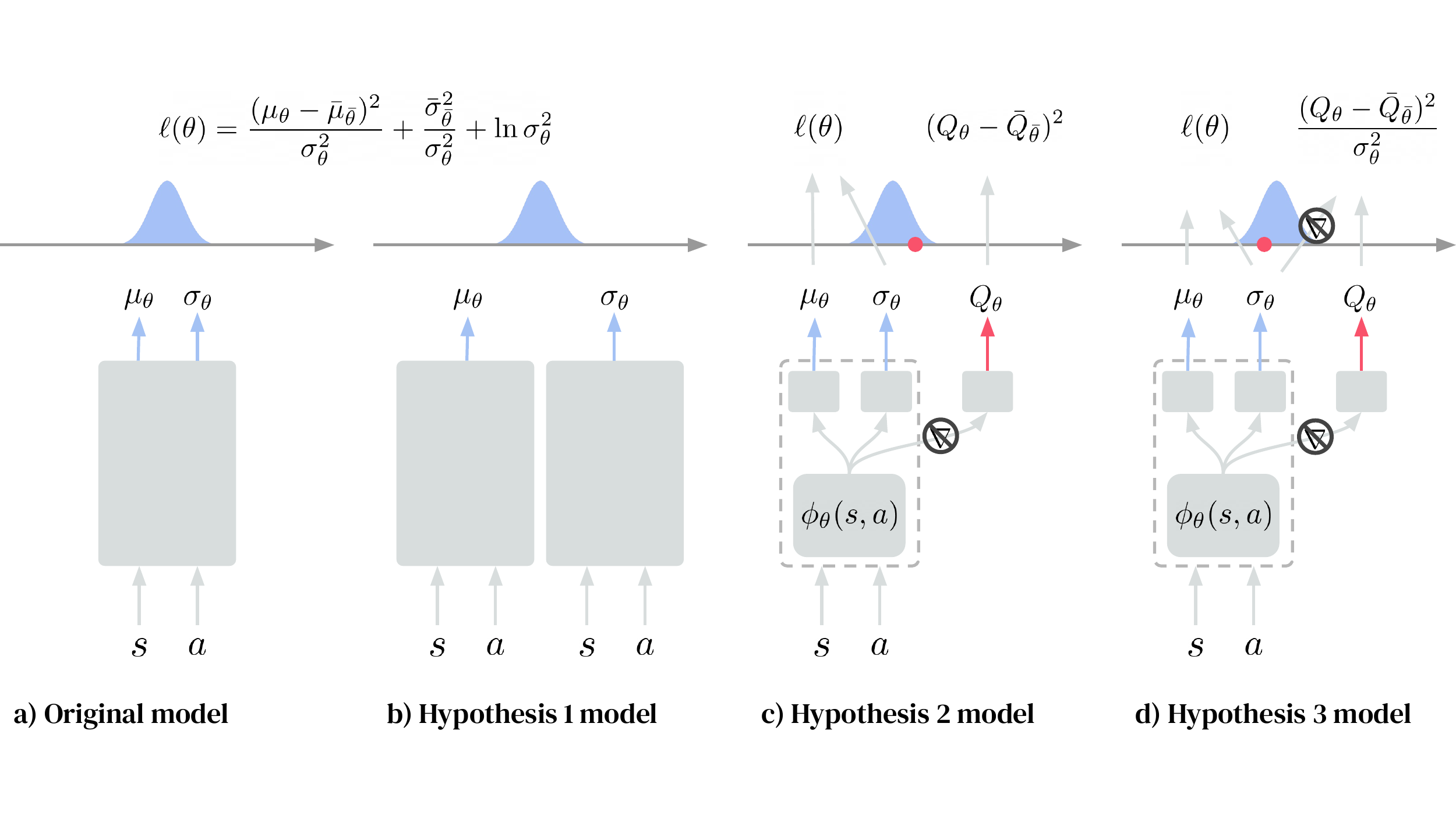}
    \caption{
        Diagrams of how neural models are used in this work.
        \textbf{(a)} The original model outputting a single Gaussian.
        \textbf{(b--d)} Models used to test Hypotheses 1 to 3, respectively.
        Hypotheses 2 and 3 use a different loss for their $Q_\theta$ heads.
    }
    \label{fig:hypothesis-diagrams}
\end{figure}

In this section we wish to deconstruct and understand what about this distributional loss
is providing such an improvement in performance over its non-distributional alternative.
In order to study more closely the cross-entropy loss, in this section we consider a 
distributional critic consisting of a single Gaussian component 
$Z_{\theta}^{(s, a)} = \mathcal{N}(\mu_\theta(s, a), \sigma^2_\theta(s,a))$.
In this special case, the cross-entropy between the target distribution and the online 
network's distributional estimate can be expressed analytically.
In an effort to lighten the notation we will consider a single element of the minibatch of 
transitions $(s, a, r, s')$ and a single sampled action $a'$ with which we
can write the distributional loss as
\begin{align}
    \ell(\theta) = H\left[ r+\gamma Z_{\bar\theta}^{(s',a')}, Z_{\theta}^{(s, a)} \right]
    &\propto
        \frac{(\mu_\theta - \bar\mu_{\bar\theta})^2}{\sigma^2_\theta}
        + \frac{\bar\sigma_{\bar\theta}^2}{\sigma^2_\theta} + \ln\sigma^2_\theta
    \intertext{where we have dropped the dependence of the Gaussian components on the state-action pair inputs. We have also
    introduced $\bar\mu_{\bar\theta}$ and $\bar\sigma_{\bar\theta}^2$ as the transformed mean and variance of the target distribution.
    This is possible due to the fact that a linear transformation of a Gaussian remains Gaussian, however note that their
    exact values can be treated as constants (due to their independence of $\theta$). We can then further simplify the expression as}
    &=
        \frac{(Q_\theta - \bar{Q}_{\bar\theta})^2}{\sigma^2_\theta}
        + \frac{\bar\sigma^2_{\bar\theta}}{\sigma^2_\theta} + \ln\sigma^2_\theta,
\intertext{
    where we've made use of the fact that the first moments of both distributions
    correspond to their associated state-action values, e.g.\ $Q_\theta(s,a)= \mu_\theta$.
    Finally, we further decouple $Q_\theta$ into a final linear layer (with weights $w$) and
    a feature vector output by the penultimate layer, denoted $\phi_\theta(s, a)$, which yields
}
    &=
        \frac{(w^\top \phi_\theta(s, a) - \bar{Q}_{\bar\theta})^2}{\sigma^2_\theta}
        + \frac{\bar\sigma^2_{\bar\theta}}{\sigma^2_\theta} + \ln\sigma^2_\theta.
\end{align}
Note that in all our experiments thus far, the feature map we refer to as $\phi_\theta(s, a)$ has been shared
among all final heads, e.g., those that parameterized the locations, scales, and mixture logits of our MoG
distributions.
This final expression reveals the similarities and differences between the traditional squared Bellman error
(SBE) and the distributional counterpart we consider in this report. The squared Bellman error, in the same notation reads:
\begin{equation}
    \ell_{\mathrm{SBE}}(\theta) 
    = (Q_\theta(s, a) - \bar{Q}_{\bar\theta}(s, a))^2 
    = (\mu_\theta(s, a) - \bar{Q}_{\bar\theta}(s, a))^2 
    = \left(w^\top \phi_\theta(s, a) - \bar{Q}_{\bar\theta}(s, a)\right)^2 
\end{equation}
assuming a deterministic critic outputting only a prediction of the mean.
Inspecting this expression leads us to three hypotheses as to what may be responsible for the
improved performance.
For each one that follows below we devise an alteration of the original distributional model to
compare it to the non-distributional baseline. The three alterations are depicted along side the original model
in Figure~\ref{fig:hypothesis-diagrams}. Further, empirically we assess each of these hypotheses on a set of four challenging
DM control suite tasks.

\paragraph{Hypothesis 1: Adaptive Mahalanobis reweighting of losses at each $(s,a)$ is beneficial.}
This hypothesis concerns the question of whether the distributional network is able to focus on those transitions that it is
relatively certain about because smaller values of $\sigma_\theta$ associates an increased weight (compared to
the non-distributional SBE) with those transitions?
To test this hypothesis we consider an alteration to the model that uses 
same loss with entirely independent architectures for $\mu_\theta$ and $\sigma_\theta$, thus controlling
for any shared representation learning in order to assess the added value of Mahalanobis reweighting alone.

\paragraph{Hypothesis 2: The additional auxiliary loss leads to better representation learning.}
Since all but the final mean and variance heads of our network are shared between the two pathways, the question is:
are the additional $\sigma_\theta$ terms in the loss serving as a grounded auxiliary loss from which to
learn better features? To test this hypothesis
we create a copy of the $\mu_\theta$ head, denoted $Q_\theta$ in
Figure~\ref{fig:hypothesis-diagrams}~(c), which is trained without the Mahalanobis reweighting and with a stop-gradient
preventing backpropagation from this new head to influence representation learning from the joint torso.
Importantly, we then use the output of the $Q_\theta$ head for policy optimization; this ensures that the
Mahalanobis reweighting \emph{only} influences the torso's feature mapping and not the $Q$-learning used for policy
optimization.

\paragraph{Hypothesis 3: Both are necessary to recover the full benefit of the distributional loss.}
Finally, is the improved performance of the studied distributional loss due to both the adaptive reweighting and the
improved feature learning?
To test this final hypothesis we take the previous model and scale the $Q_\theta$ head's loss by $\sigma_\theta$
without propagating any additional gradients down the torso.
Once again, the only signal training the torso's feature mapping is the studied distributional loss,
and the only signal training the policy is $Q_\theta$, which is now able to adaptively reweight $(s, a)$ examples
during training.

\begin{figure}
    \centering
    \includegraphics[width=0.9\textwidth]{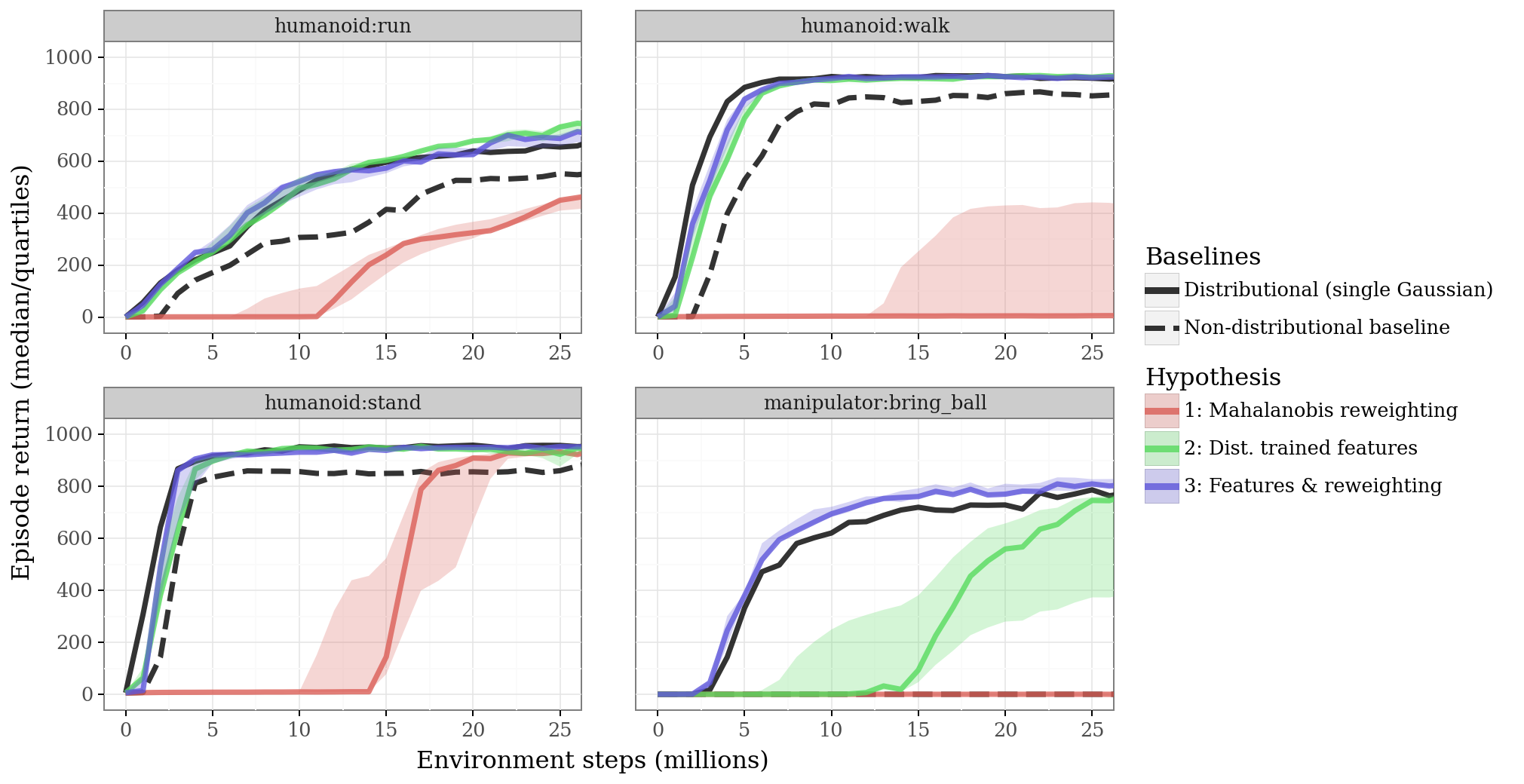}
    \caption{
        Testing our three hypotheses for which elements of the distributional loss are 
        contributing to the success of the sample-based distributional loss,
        showing median and quartiles across 5 seeds.
        Combining both the feature learning and the Mahalanobis reweighting
        recovers the performance of the fully distributional agent.
    }
    \label{fig:hypothesis-testing}
\end{figure}

Our results, presented in Figure~\ref{fig:hypothesis-testing}, show quite convincingly that both the effect on
feature learning and the adaptive reweighting of instances are responsible for the improved performance, as
evidenced by the blue Hypothesis~3 curve recovering the performance of the single Gaussian baseline in black.
In contrast, while the feature learning alone (green) is helpful in three out of four tasks, the reweighting
alone (red) can have a catastrophic effect on learning performance.

\section{Sensitivity analysis}

\subsection{Hyperparameter Sensitivity}
\label{sec:sensitivity}

\begin{figure}
    \centering
    \includegraphics[width=0.85\textwidth]{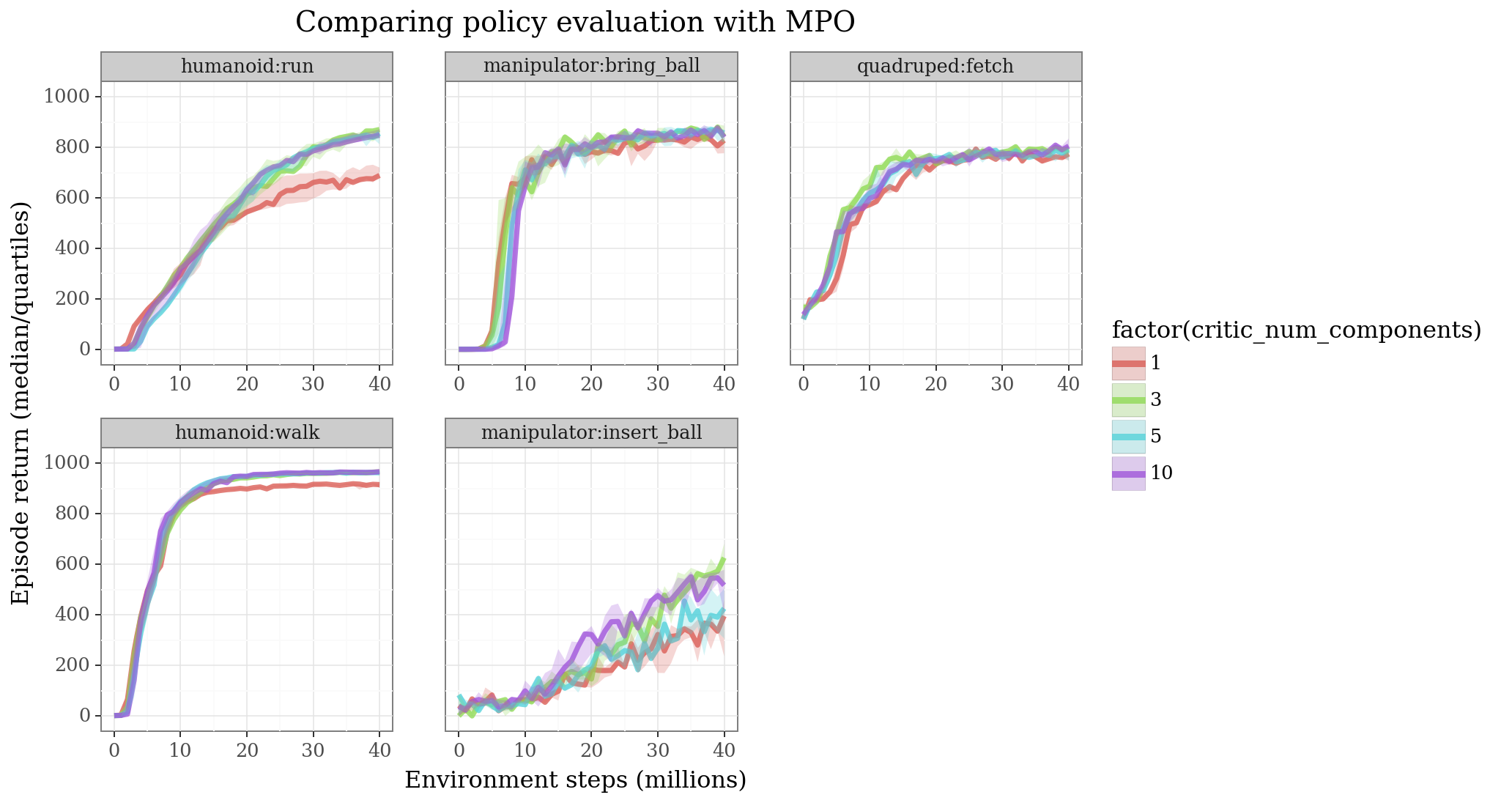}
    \caption{
    Sensitivity analysis of the number of Gaussian mixture components on the selected suite of
    challenging DeepMind control suite tasks. Solid curves are means over three seeds. Performance is
    quite robust to the number of components, though it seems more than one are needed to reach the
    state-of-the-art on the humanoid tasks.
    }
    \label{fig:sensitivity-num_components-default-nets}
\end{figure}

\begin{figure}
    \centering
    \includegraphics[width=0.8\textwidth]{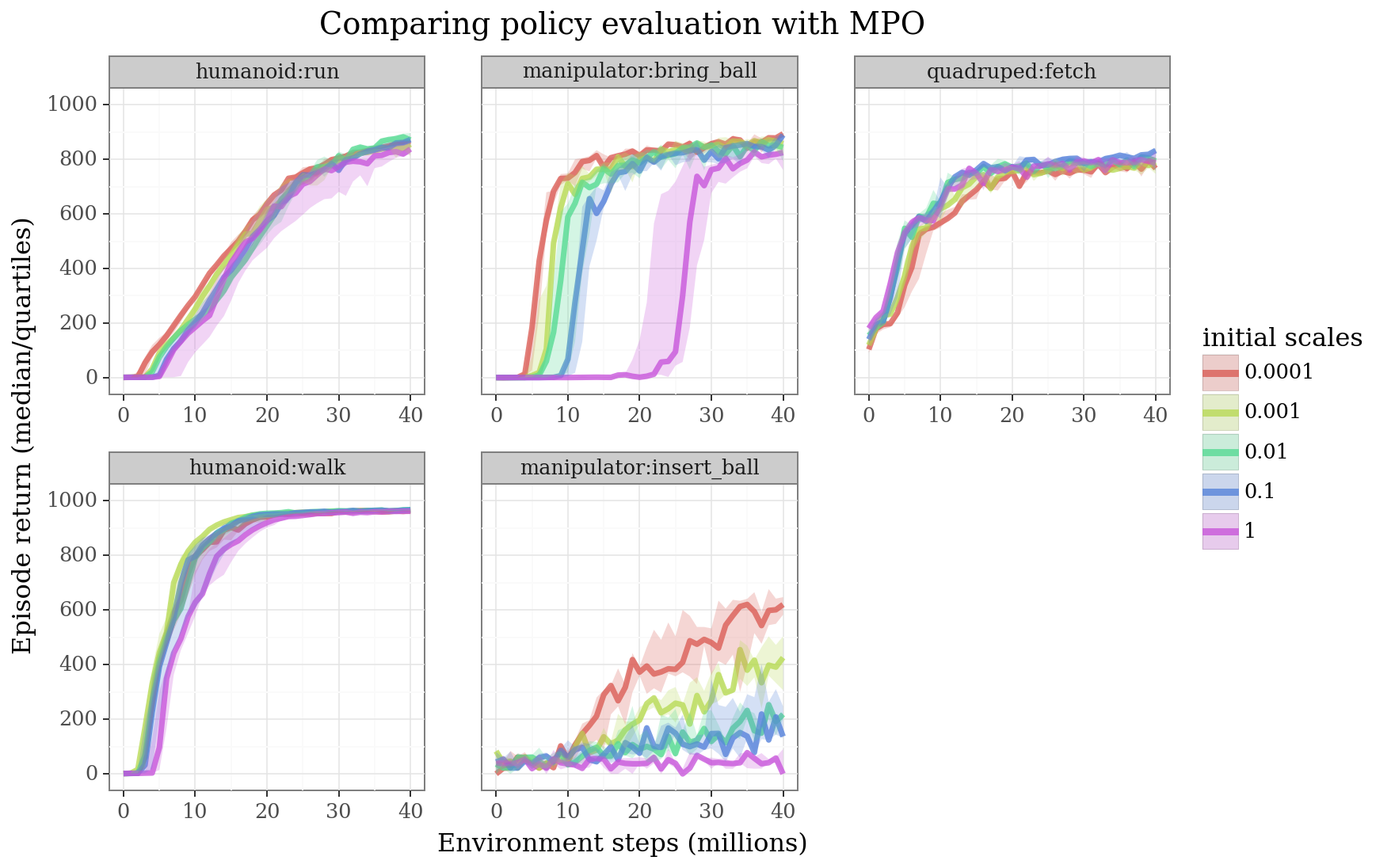}
    \caption{
    Sensitivity analysis of the initial scale of Gaussian mixture components on the selected suite of
    challenging DeepMind control suite tasks. The number of components was set to the default of 5 for
    this comparison. Solid curves are means over three seeds. Performance is relatively robust to the
    initial scale of components, though on manipulator tasks it seems important to initialize the scales
    to a very small number.
    }
    \label{fig:sensitivity-initial_scales-default-nets}
\end{figure}

The mixture-of-Gaussians parameterization of the distributional critic studied in this report does
introduce new hyperparameters of its own. While we do not need to specify the \texttt{vmin/vmax}
and \texttt{num\_atoms} hyperparameters as in the categorical parameterization, we do specify the
number, initial scale, and initial location of the mixture components. In all experiments we set
the initial locations of all components to the origin, however we did carry out a sensitivity
analysis on the number and initial scale of the Gaussian components.
While performance seemed to be quite robust to the number of components as seen in
Figure~\ref{fig:sensitivity-num_components-default-nets}, it did seem to be somewhat sensitive to the
choice of initial scale on the manipulator tasks.
Indeed, Figure~\ref{fig:sensitivity-initial_scales-default-nets} shows performance on manipulator
favouring small values of initial scale over those that are closer to unity.

\subsection{Generalization to other policy optimization methods}

Although this work focuses on the MPO algorithm for policy optimization, 
our results also suggest that the benefits discussed in this report generalize to
other approaches that make use of a critic (e.g.\ DDPG). Recall that DDPG
with a C51 distributional critic is the very competitive agent known as D4PG~\citep{d4pg}.

\section{Discussion and future work}

We have shown that fitting more distributional moments has lead to better performance; and
our analysis strongly suggests that this enhancement is due to better representation learning.
The question remains as to what makes this representation better. Could it be that the
losses due to the additional moment has a regularizing effect on feature learning, making
it more robust to time-varying distributions of returns and/or state-action pairs?

Our cross-entropy loss has been shown not to be a norm under which our distributional Bellman
operator is a contraction. While this was a simple first attempt with positive empirical
results, we would like to investigate ways of deriving a similarly simple approach from a loss
that would guarantee contraction. For example, in the single Gaussian case, we could try the
2-Wasserstein distance, or explore the Cram\'er
distance~\citep{bellemare2017cramer, rowland2018analysis} as a general solution.

\section*{Acknowledgments}
We would like to thank Yannis Assael for translating our original Tensorflow implementation to JAX.

\bibliographystyle{abbrvnat}
\bibliography{references}

\end{document}